\documentclass[letterpaper, 10 pt, conference]{ieeeconf}  

\pdfoutput=1

\IEEEoverridecommandlockouts                              

\usepackage{amsmath, amsthm, amsfonts, amssymb,color}
\usepackage[ruled,vlined,noend]{algorithm2e}
\usepackage{blindtext}
\usepackage{graphicx}
\usepackage[utf8]{inputenc}
\usepackage{mathtools}
\usepackage{tikz, cite}
\usepackage[font=footnotesize,labelformat=simple]{subcaption}
\usepackage{tabu}

\newtheorem{example}{Example}
\usepackage{epsfig} 
\usepackage{psfrag}
\usepackage{mathrsfs}

\usepackage{multirow}
\usepackage{array,booktabs,arydshln,xcolor}
\usepackage{soul}
\usepackage{mathtools}
\usepackage{nccmath}
\usepackage{caption}
\usepackage{nccmath}
\usepackage{comment}
\usepackage{pdfpages}
\usepackage{comment}
\renewcommand{\vec}[1]{\boldsymbol{#1}}
\usepackage{float}
\usepackage{graphicx}
\usepackage{epstopdf}
\usepackage{color}
\usepackage{verbatim}
\usepackage{tikz,times}
\usepackage{xcolor}
\usepackage{lipsum}
\usepackage{makecell}
\let\labelindent\relax

\usepackage{enumitem}
\usepackage{multirow}

\usepackage{stfloats}
\usepackage{siunitx}

\usepackage{xspace}

\makeatletter
\DeclareRobustCommand\onedot{\futurelet\@let@token\@onedot}
\def\@onedot{\ifx\@let@token.\else.\null\fi\xspace}

\def\eg{\emph{e.g}\onedot} 
\def\ie{\emph{i.e}\onedot}

\makeatother

\title{\LARGE \bf
Closing the Loop: Motion Prediction Models beyond Open-Loop Benchmarks 
}

\author{Mohamed-Khalil Bouzidi$^{1,3}$,
Christian Schlauch$^{2,3}$, 
Nicole Scheuerer$^{3}$, 
Yue Yao$^{1,3}$,\\ 
Nadja Klein$^{2}$,
Daniel Göhring$^{1}$,
Jörg Reichardt$^{3}$
\thanks{$^{1}$  Freie Universität Berlin, Germany }
\thanks{{\tt\small \{firstname.lastname@fu-berlin.de\}}}
\thanks{$^{2}$ Karlsruhe Institute of Technology, Germany }
\thanks{{\tt\small \{firstname.lastname@kit.edu\}}}
\thanks{$^{3}$ Continental AG}
\thanks{{\tt\small \{firstname.lastname@continental.com\}}}
\thanks{This work is funded by the German Federal Ministry for Economic Affairs and Climate Action within the project ''NXT GEN AI METHODS''.}
}

\begin{document}

\maketitle
\thispagestyle{empty}
\pagestyle{empty}

\begin{abstract}
Fueled by motion prediction competitions and benchmarks, recent years have seen the emergence of increasingly large learning-based prediction models, many with millions of parameters, focused on improving open-loop prediction accuracy by mere centimeters. However, these benchmarks fail to assess whether such improvements translate to better performance when integrated into an autonomous driving stack. In this work, we systematically evaluate the interplay between state-of-the-art motion predictors and motion planners. Our results show that higher open-loop accuracy does not always correlate with better closed-loop driving behavior and that other factors, such as temporal consistency of predictions and planner compatibility, also play a critical role. Furthermore, we investigate downsized variants of these models, and, surprisingly, find that in some cases models with up to 86\% fewer parameters yield comparable or even superior closed-loop driving performance.
Our code is available at https://github.com/aumovio/pred2plan.
\end{abstract}
\section{Introduction} \label{sec:intro}

Autonomous driving (AD) is widely considered a cornerstone technology for future mobility. Its adaptation and acceptance will depend on how comfortable and safe autonomous vehicles (AVs) will be. Given the inherent difficulty of the task, many practical AD systems are based on modular architectures that decompose the problem into subtasks such as perception, motion prediction, and planning \cite{hagedorn_integration}. In these architectures the perception module estimates the static and dynamic state of the environment, upon which the prediction module forecasts the future behavior of surrounding traffic participants (TPs). These predictions then serve as input to the planning module, which generates a feasible, safe, and comfortable trajectory for the AV. While recent advances have also driven progress in end-to-end learning approaches \cite{navsim}, modular systems remain popular due to advantages in interpretability, safety validation, and developmental tractability.

As a result, a large body of research has focused on improving single modules, for example the prediction module, in isolation. Catalyzed by the recent availability of large-scale data sets of recorded traffic scenarios and associated competitions such as nuScenes\cite{caesar_nuscenes_2020} , Argoverse\cite{wilson_argoverse2_2021}, and Waymo Open \cite{ettinger_waymo_2021}, many different learning-based approaches were developed that only aim to predict the evolution of a traffic scenario based on a given observed history and road layout over several seconds, \ie, in an open-loop (OL) setting, without ever acting on these predictions with downstream planning modules. These benchmarks typically rank models based on how close they are to the ground truth trajectory, \ie, how accurate the predictions are. Over time, this OL evaluation paradigm has led to a competitive push toward increasingly large and complex models, aimed at shaving off mere centimeters of prediction error to climb leaderboard rankings reporting prediction accuracies far beyond what humans are typically capable of \cite{wirth2019}. For instance, in the Argoverse 2 leaderboard, prediction error between top-100 entries increases by less than \SI{15}{cm}, while on Waymo only \SI{8}{cm} separate the top 20 entries — despite model size differences spanning millions of parameters. 

These trends raise a critical question:
Do marginal improvements in open-loop prediction accuracy translate to meaningful gains in closed-loop (CL) driving performance? Or, conversely, do we observe diminishing returns, where increased model complexity fails to improve, or even harms, actual vehicle behavior when integrated into a full system? This is especially important as any increasing model size is paid for by increased compute effort and hardware requirements.
Another important question is whether other criteria become more relevant, and what challenges emerge, when SotA predictors are actually integrated with a planning module in an AV. In other words,  what constitutes a good predictor when paired with a planner.  What makes a good planner robust if predictions are sub-optimal? From a system perspective, it is the highly dynamic interplay between prediction and planning algorithms that ultimately determines driving performance. 

With this work, we contribute towards a more realistic and holistic evaluation of prediction methods in the context of motion planning from system perspective. Specifically, we:
\begin{itemize}[leftmargin=*]
 \item  introduce a closed-loop simulation and evaluation framework to test several SotA prediction models together with different planner based on nuPlan \cite{nuplan} and UniTraj\cite{unitraj}
\item investigate the interplay between planning and prediction algorithms qualitatively
\item elucidate the relation between open-loop prediction accuracy and closed-loop driving performance quantitatively
\item examine downsized variants of SotA predictor models to assess whether large model sizes and the corresponding gains in OL accuracy yield any benefit to CL-driving.
\end{itemize}

The paper is organized as follows: 
We first discuss typical evaluation metrics for open-loop prediction algorithms and their relevance for the planning task, as well as metrics for planning performance. We then introduce our closed-loop evaluation pipeline and its individual components, the different datasets used for evaluation, the different prediction and planning algorithms and their variants under test. Finally, we show and discuss a number of quantitative experiments that illuminate the complex interplay between open-loop prediction accuracy and closed-loop driving performance.

\section{Background and Related Work} \label{sec:related_work}

\subsection{Problem Definition Motion Prediction}

Motion predictors aim to estimate the states $x_{i,t|1},..,x_{i,t|N}$ of a target agent $i$ over a sequence of $N$ time steps into the future, given a scene context as input. This scene context $C_t= \{ X_{t-H}, \dots, X_t, M_t, G_t \}$ consists of the observed states $X_t = \{x_{i, t}, x_{j\neq i,t}\}$ of the target $i$ and its surrounding agents $j\neq i$ in the set of all agents $\mathcal{A}$ in the scene over a history length $H$, as well as the road's topology $M_t$ (\eg, drivable area, lane markings) and dynamic traffic guidance $G_t$ (\eg, traffic lights or speed limits).

A target's predicted motion can be described by the conditional distribution
    $p(x_{i,t|1}, \dots, x_{i,t|N} | C_t)$ 
which is generally multi-modal reflecting a multitude of possible maneuvers and the interactions with the other agents in the scene. Typically, predictions are modeled as Gaussian Mixture Models (GMMs) where each mixture component $m$, \ie, \emph{mode}, represents a distinct predicted trajectory $\hat{x}^m_{i,t|k}, k\in {1,..,N}$.

\subsection{Open-Loop Evaluation of Motion Predictors}

Open-loop testing evaluates the prediction module as a standalone component independent of the planning module, operating under the assumption that all improvements in prediction accuracy will automatically lead to better planning outcomes. Recent large-scale motion data sets and their associated motion prediction competitions such as nuScenes\cite{caesar_nuscenes_2020} , Argoverse\cite{wilson_argoverse2_2021}, and Waymo Open (WOMD) \cite{ettinger_waymo_2021} have fostered this paradigm, providing standardized open-loop evaluation frameworks for assessing prediction models. 

The most common benchmark metrics are the minimum Average Displacement Error ($\text{minADE}_{K}$) and minimum Final Displacement Error ($\text{minFDE}_{K}$). The metric $\text{minADE}_{K}$ calculates the Euclidean distance in meters between the ground-truth trajectory and the best of $K$ outputted predictions as an average of all future time steps (standard are K=1 and K=6).  
In contrast, $\text{minFDE}_{K}$ focuses solely on the prediction error in the final time step. 


OL benchmarks typically neglect the probabilistic nature of predictions since the predicted entropies and probabilities of the GMM modes have only an indirect effect on the displacement error metrics. However, these estimates carry important information about the models uncertainty and  which can be crucial for planners \cite{khaled}. Prior work introduced metrics such as the $\text{minNLL}_{K}$ which we also incorporate in our evaluation to account for the uncertainty in the predictions \cite{ivanovic2019trajectron, schlauch2023informed}. The $\text{minNLL}_{K}$ measures the negative log likelihood of ground truth trajectory in the best of $K$ modes of the GMM as an average of all future time steps. 

Besides, recent research highlights the significant distribution shift between different datasets \cite{ unitraj, yao_improving_2024}. Consequently, out-of-distribution testing across datasets has been proposed as a more robust approach to evaluate model generalization.

It is important to note that the motion prediction challenges only require a single prediction per agent and scenario \ie, only for one single timestep. This is in contrast to the situation in the vehicle where predictions are made continuously while driving. Prediction challenges presently do not account for temporal consistency which is one of the major motivations for our work. 

\subsection{Problem Definition Motion Planning}

The objective of a planner is to generate a safe and comfortable trajectory along the reference path. This can be formulated as a finite-horizon optimal control problem:

\begin{equation} 
\begin{aligned}
\nonumber
 &\underset{\vec{z}_{k},\vec{u}_{k} \in [0,N]}{\min} \quad  \sum_{k=0}^{N-1} J_k(\vec{z}_k, \vec{u}_k) + J_N(\vec{z}_N) \\
\text{s.t.} \quad & \vec{z}_{k+1} = f(\vec{z}_{k},\vec{u}_{k}),\ \vec{z}_0 = \vec{z}(0),\ \vec{z}_{k} \in \mathcal{Z},\ \vec{u}_{k} \in \mathcal{U}
\end{aligned}
\end{equation}

Here, \( J_k \) denotes the running cost at time step \( k \), \( J_N \) is the terminal cost. The state space model with states $\vec{z}$ and control inputs $\vec{u}$  approximates the vehicle physics, \eg by a discretized kinematic bicycle model.
The constraint-sets \( \mathcal{Z} \) and \( \mathcal{U} \) represent the feasible state and input spaces, respectively, and can contain actuator and dynamic limitations, traffic rule adherence, and collision avoidance constraints.
For collision avoidance constraints, in modular architectures, the planner incorporates predicted trajectories of surrounding traffic participants, as provided by the motion prediction models. 

\subsection{Closed-Loop Evaluation of Motion Planner}

Motion planning approaches have traditionally been evaluated in CL benchmarks to assess their ability to ensure safety and passenger comfort by controlling the autonomous vehicle in the environment.
Among notable benchmarks, CommonRoad \cite{althoff_commonroad_2017} was one of the earliest. More recently, benchmarks such as CARLA Leaderboard \cite{dosovitskiy_carla_2017}, with a synthetic simulation environment, and nuPlan \cite{nuplan} offering extensive real-world driving logs, have gained prominence for evaluating learning-based planners.

In these benchmarks, a range of metrics is provided to evaluate performance, including at-fault collision rate (CR), driving area compliance, progress along the reference path, and time-to-collision (TTC). Comfort is also assessed through measures such as lateral and longitudinal jerk, accelerations, yaw rate, and yaw acceleration. These metrics are typically combined into an overall score. For example, nuPlan calculates a score between $0$ and $1$, where $1$ indicates optimal performance and $0$ the worst. This score - which is also used in our evaluation - partially weighs the aforementioned metrics, while others, such as at-fault collision and drivable area compliance, act as hard constraints and result in a scenario score of $0$ if violated.

While the rise of learning-based planners in recent years has popularized the use of open-loop metrics similar to those used in motion prediction, several recent works \cite{pmlr-v229-dauner23a, navsim, interplan} have emphasized the limitations of open-loop evaluation. These works demonstrated that open-loop performance of planners fails to reflect the true closed-loop performance of autonomous vehicle by leveraging the nuPlan framework and primarily focusing on end-to-end and learning-based planners. 
In contrast, our work addresses the misconception regarding open-loop evaluation specifically within the distinct task of motion prediction, which differs fundamentally from planning. We consider a modular approach, where a learning-based motion predictor provides predictions to optimization-based planners. Although such modular approaches are widely established in both academia and industry \cite{hagedorn_integration, ziegler2014, schwarting2018,fan2018baiduapolloemmotion,bouzidi2025reachabilitybasedcontingencyplanningmultimodal, chen_interactive_2021, Cui_2021_ICCV}, they are rarely assessed in the aforementioned works.

\section{Framework} \label{sec:framework}

\begin{figure*}[t]
\vspace{0.6em} 
    \centering
    \includegraphics[trim = 0mm 0mm 0mm 0mm, width=1\textwidth]{"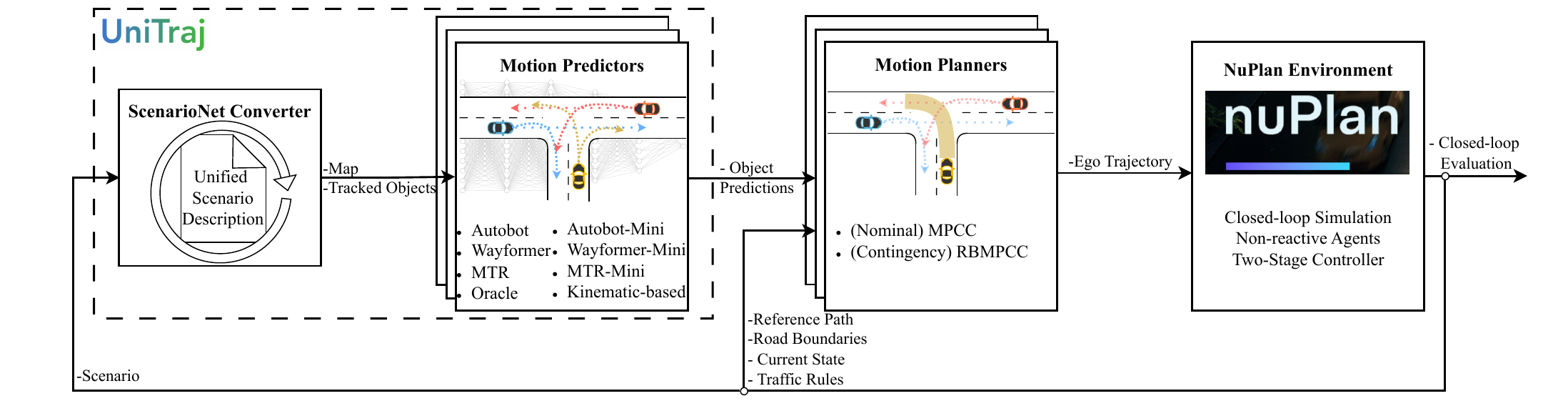"}
    \caption{Our closed-loop evaluation framework integrates and tests different SotA prediction models on different planners}
    \label{fig:method}
    \vspace{-1.2em} 
\end{figure*}

Our closed-loop simulation and evaluation framework, illustrated in Fig. \ref{fig:method}, combines two previously independent systems: UniTraj \cite{unitraj}, which provides a unified data interface for motion predictors, and nuPlan \cite{nuplan}, a large-scale motion planning benchmark platform. This integration enables us to systematically simulate and evaluate the performance of various combinations of state-of-the-art (SotA) motion predictors and planners in a realistic closed-loop autonomous driving setup. These predictor-planner combinations are executed within nuPlan, allowing us to observe and compare the autonomous vehicle's behavior under different configurations.

\subsection{UniTraj Prediction Framework} \label{sec:unitraj}

Cross-dataset training and testing can be crucial for scaling models and evaluating their generalization. However, motion datasets are published in various data formats with individual sampling frequencies, semantic labels and associated software stacks. Maintaining multiple dataloaders for each dataset for every motion predictor is a significant engineering burden. Arbitrary design decisions in the dataloading also complicate scientific comparisons between model designs. The UniTraj framework addresses these issues by unifying selected datasets into a common format using the ScenarioNet \cite{li_scenarionet_2023} converter. The ScenarioNet converter maps semantic labels of agents and map elements to the Waymo Open protocols \cite{ettinger_waymo_2021} and interpolates observations to a common sampling frequency. UniTraj offers a unified data interface for motion predictors, providing agent states (position, velocity, heading, size) as well as static and dynamic map elements (centerlines, road boundaries, crosswalks, obstacles, traffic light states) for selected timeframes. To facilitate cross-dataset training and open-loop evaluation, we streamline this framework and integrate the Shifts dataset \cite{malinin2021shifts} converter in addition to the provided NuScenes \cite{caesar_nuscenes_2020}, Argoverse2 \cite{wilson_argoverse2_2021}, Waymo Open \cite{ettinger_waymo_2021} and nuPlan dataset \cite{nuplan} converters. These datasets are recorded and labeled using different instruments and cover a broad range of geographical locations, reflecting real-world distribution shifts.

\subsection{nuPlan Simulation Framework} \label{subsec:nuplan}
We use the nuPlan simulator for our experiments, as it provides a realistic and large-scale simulation environment built on over \SI{1300} hours of real-world driving data. Each scenario is initialized from a real-world log and includes a high-definition semantic map, with a reference path, drivable area for the ego vehicle, and relevant traffic rules such as local speed limits. These elements can be accessed to initialize and constrain the planner.

At every timestep, the simulation can provide the current history of the ego vehicle and surrounding agents (vehicles, pedestrians, bicycles) in the form of box-level observations, with states such as pose and velocity, as typically output by a perception module. The behavior of traffic participants can either be replayed from the logged data (non-reactive agents) or simulated as reactive agents using a driver model, such as the Intelligent Driver Model (IDM) \cite{treiber_congested_2000}. In this work, we use \emph{non-reactive} agents to ensure better comparability and isolate the effects of different motion predictors, as agent behavior remains consistent across simulation runs. 

Finally, the trajectory output by the planner in every timestep is passed to a built-in Linear Quadratic Regulator (LQR) within nuPlan, to control the AV which is simulated by the kinematic bicycle model. This setting called two-stage control architecture ensures that the ego vehicle cannot follow physically infeasible trajectories.

\section{Methods} \label{sec:methods}

For our closed-loop evaluation, we incorporate several SotA motion prediction models, along with their variants that differ in model capacity (\ie, reduced parameter count). Each predictor is paired with two different motion planning modules to form distinct predictor-planner combinations.
This section summarizes the specific motion predictors and planners used in our experiments, detailing their differences  and the rationale behind their selection.

\subsection{Motion Predictor} \label{sec:pred}

We analyze variants of three SotA deep learning predictors, namely Autobot \cite{girgis2022latent}, Wayformer \cite{nayakanti_wayformer_2022} and MotionTransformer \cite{shi_motion_2022}, in addition to kinematic predictors and an oracle predictor. The selected deep learning predictors are transformer-based architectures that factorize the distribution over the target's future states as mixtures of bivariate Gaussians at each timestep. They show competitive performance on the previously mentioned public OL benchmarks.

\textbf{MotionTransformer (MTR)} \cite{shi_motion_2022} employs a transformer-based encoder-decoder architecture and  jointly optimizes global intention localization and local movement refinement. We also analyze a downscaled \textbf{MTR-Mini} version by halving the width and number of layers of the published design leading to a decrease of 86\% in the number of parameters.

\textbf{Wayformer} \cite{nayakanti_wayformer_2022} consists of a transformer-based early-fusion encoder with a transformer-based decoder using compute-efficient latent query encodings. The loss is defined on the best mode in the estimated mixtures. 
In addition, we introduce a \textbf{Wayformer-Mini} version which halves the width of layers, decreasing the parameter count by 74\%.

\textbf{Autobot} \cite{girgis2022latent} is a lightweight transformer-based encoder-decoder architecture, that interleaves the scene-context during encoding to learn equivariant features. The model uses a probabilistic loss on the complete estimated mixtures. Since Autobot's layer width is already small, we reduce its model capacity by halving the number of layers instead, denoted as \textbf{Autobot-Mini},  decreasing parameter count by 50\% .

\textbf{Kinematic Predictors} predict the future TP trajectories based only on kinematics, ignoring the scene context. We employ a simple constant velocity model along the current TP's heading. 
Additionally, to match the 6‐mode output of our other predictors, we employ a multi‐model kinematic predictor that spawns 6 hypotheses by perturbing the agent’s yaw rate (zero, small left, small right) or acceleration (zero, moderate acceleration, moderate and strong braking).

\textbf{Oracle Predictor} serves as an upper bound on the open-loop prediction performance by replaying the observed ground truth motion of the surrounding traffic participants.
\vspace{-.5em} 
\begin{table}[!htb]
\scriptsize
\caption{Model Capacity Comparisons}
\vspace{-10pt}
\label{tab:splits}
\begin{center}
\resizebox{0.9\linewidth}{!}{
\begin{tabular}{l*{2}{r}}
\toprule
\textbf{Model} &  \# Parameters (in Mio) & Forward Pass (in GFlops)  \\
\midrule

MTR & 65.2 & 10.34 \\
MTR-Mini & 9.3 &  1.66 \\
Wayformer & 15.2 &  6.72 \\
Wayformer-Mini & 3.9 &  2.15 \\
Autobot & 1.5  &  1.91 \\
Autobot-Mini & 0.74 & 1.16  \\
\bottomrule
\end{tabular}}
\end{center}
\vspace{-0.5pt}
\end{table}

\subsection{Motion Planner} \label{sec:planner}

To investigate how the choice of planner shapes the influence of the motion predictor on AV behavior, we select two planning approaches, each representing a distinct and established class of methods in terms of how they handle multi-modal TP predictions.

\textbf{Model Predictive Contouring Control (MPCC) \cite{brito1}} falls into the category of single trajectory planners \cite{ziegler2014, schwarting2018, fan2018baiduapolloemmotion}, which use only the single most probable predicted trajectory of surrounding agents for decision-making, while discarding all other modes. This approach is widely adopted for its simplicity and computational efficiency, relying on frequent replanning to address prediction uncertainty.
The planner maximizes progress along a given reference path while minimizing lateral deviation from it. Smoothness is achieved by penalizing high jerk and steering angle rates. To enhance safety, a potential field term extends the basic formulation encouraging to maintain a safe distance from the predicted TP trajectories.

\textbf{Reachability-based Branch Model Predictive Contouring Control (RBMPCC)\cite{bouzidi2025reachabilitybasedcontingencyplanningmultimodal}}  belongs to the class of contingency planners \cite{khaled, chen_interactive_2021, Cui_2021_ICCV}, which  account for multi-modal agent predictions by maintaining multiple planning strategies, each associated with different prediction modes. To delay commitment under uncertainty, all strategies are constrained to be identical during the initial timesteps, enabling decision postponing until the uncertainty resolves. Our selected implementation extends the conventional MPCC by constructing a scenario tree based on reachability analysis. This leads to the extraction of different ego-vehicle driving corridors,  that represent individual branches in the tree, where each branch contains a dedicated contingency plan avoiding collisions with one or more predicted modes.

\section{Evaluation} \label{sec:results}

\begin{figure*}[!htb]
\vspace{0.6em} 
    \centering
    \includegraphics[trim = 0mm 0mm 0mm 0mm, width=1\textwidth]{"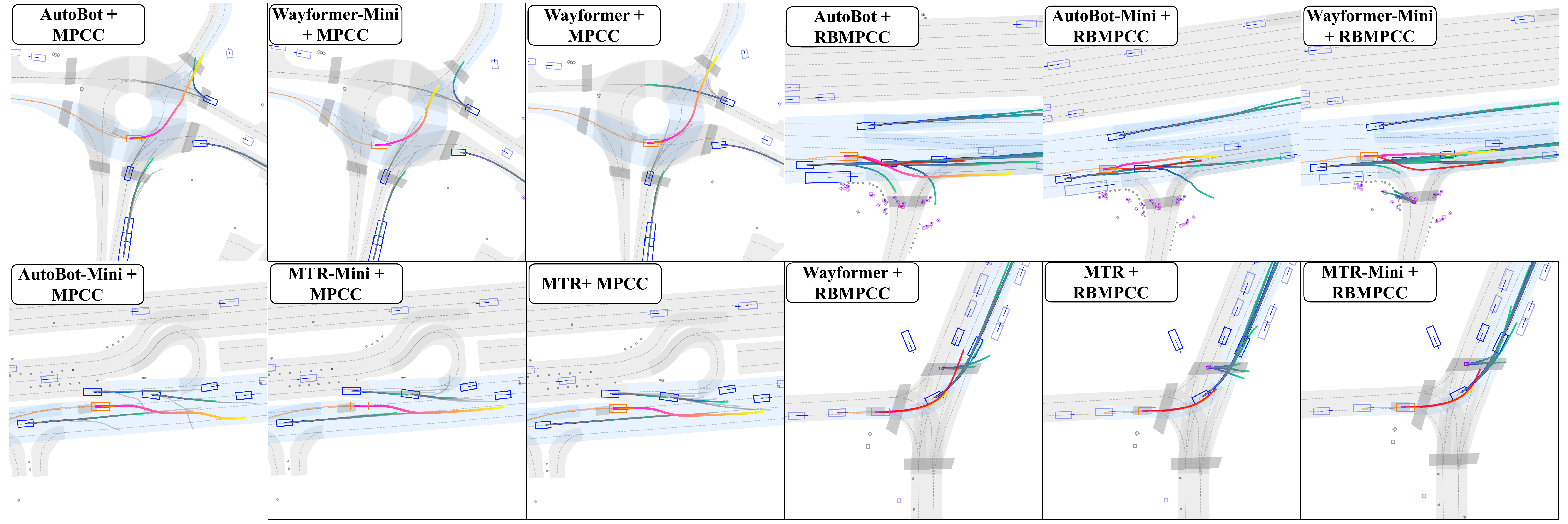"}
    \caption{Different examples of scenarios that we run to compare different combinations of predictor and planner}
    \label{fig:examples}
    \vspace{-1.2em} 
\end{figure*}

\subsection{Experimental Setup} \label{sec:exp}

\textbf{Open-Loop Predictor Training and Testing:}
To ensure a fair comparison, we configure the UniTraj dataloader equally for all models. Each scene is of \SI{7}{\second} length (\SI{1}{\second} past, \SI{6}{\second} prediction horizon) and sampled at \SI{10}{Hz}. They are centered, rotated, and cropped around the target's current state and include the states of the closest surrounding agents as well as the map elements in a \SI{100}{\meter} radius. We use NuScenes, Argoverse2, Waymo Open and Shifts for training and in-distribution (ID) OL evaluation.
We do not train on nuPlan, but test on the nuPlan validation split to enable out-of-distribution (OOD) OL and CL evaluation as a reflection of distribution shifts in real-world applications. Since nuPlan does not provide preselected prediction targets as the other datasets, we filter out trivial targets by employing a threshold on Kalman difficulty proposed in \cite{unitraj}, yielding \SI{154112} test samples for OL OOD evaluation.

\textbf{Filtering Interactive Scenarios for Simulation:}
It has been previously observed in  \cite{interplan, navsim} that the majority of scenarios in the nuPlan dataset, and in many AD datasets in general, are relatively trivial, such as motion at constant speed or stationary conditions with minimal interaction between agents.
However, such cases provide limited insight into the performance of motion predictors. Instead, we focus on long-tail scenarios where anticipating TP behavior is critical for collision avoidance and ride comfort.
To identify such interactive situations, we employ a scenario filtering method called Minimum Time-to-Conflict Point Difference ($\Delta \text{TTCP}_{\min}$),  applied in \cite{zhan_interaction_2019}. In contrast to Kalman difficulty, with this criterion we consider relevance from the AV’s perspective.
It first detects whether two agents (in our case the ego vehicle and another TP) have a conflict point, \ie, a spatial location where their paths intersect or merge.
For the conflict point, we then calculate:
\begin{equation}
\Delta \text{TTCP}_{\min} = \min_{t \in [T_{\text{start}}, T_{\text{end}}]} \left| \frac{d_{1,t}}{v_{1,t}} - \frac{d_{2,t}}{v_{2,t}} \right|
\end{equation}
where \( d_{i,t} \) and \( v_{i,t} \) denote the distance to the conflict point and the velocity of agent \( i \) at time \( t \). A small $\Delta \text{TTCP}_{\min} \leq 3  $ indicates that at a point in time, agents needed to interact to avoid arriving at the conflict point almost at the same time.
With this criterion, we extracted 50 diverse interactive scenes  (\SI{15}{s} length, sampling time \SI{0.1}{s}) from the nuPlan validation set, such as lane changes, merging, pedestrian crossings, and intersections.

\subsection{Results}\label{sec:res}

\begin{table*}[b]
    \vspace{-0.5em} 
    \caption{OL performance of predictors ID vs. OOD in nuPlan vs. over the nuPlan simulation (SIM) for every timestep. Additionally we show the Temporal Consistency score $TC^{SIM}$ of the simulation and the average inference time $t^{SIM}_i$.  For the kinematic-based predictor, left value is constant velocity, right the multi-model predictor. See text for details.}
    \vspace{-0.6em} 
    \centering
    \begin{tabular}{|p{2.05cm}||m{1.5cm}|m{1.5cm}|m{1.6cm}|m{1.6cm}|m{1.5cm}|m{1.5cm}|m{1.3cm}|m{1.3cm}|}
    \hline
    \rule{0pt}{9pt} 
     & \textit{minADE}$^\mathrm{ID}_6\downarrow$ 
     & \textit{minFDE}$^\mathrm{ID}_6\downarrow$ 
     & \textit{minADE}$^\mathrm{OOD}_6\downarrow$ 
     & \textit{minFDE}$^\mathrm{OOD}_6\downarrow$ 
     & \textit{minADE}$^\mathrm{SIM}_6\downarrow$ 
     & \textit{minFDE}$^\mathrm{SIM}_6\downarrow$ 
     & \textit{TC}$^\mathrm{SIM}\downarrow$ 
     & $\bar{t}^\mathrm{SIM}_i\,[\mathrm{ms}]\downarrow$ \\ 
    \hline
    \textbf{Oracle}    & 0 & 0  & 0  & 0 & 0 & 0 & 0&  0 \\ 
    \hline
    \textbf{MTR}                 & 0.60  & 1.31& 0.64 &1.43 & 0.94 & 2.20 & 2.61 &60\\ 
    \hline
    \textbf{MTR-Mini}            & 0.64 & 1.40  & 0.64  & 1.43 &  0.98& 2.31& 2.82 &  37 \\ 
    \hline
    \textbf{Wayformer}           &  0.50& 1.07  & 0.57  & 1.23&  0.87 & 1.95&  3.18 & 21 \\ 
    \hline
    \textbf{Wayformer-Mini}      & 0.53 & 1.15 & 0.57 & 1.23 & 0.88 & 1.99 & 2.70 & 20 \\ 
    \hline
    \textbf{Autobot}             & 0.60 & 1.27  & 0.69 & 1.46 & 1.07 & 2.47 & 5.46& 21 \\ 
    \hline
    \textbf{Autobot-mini}        & 1.20 & 3.20 & 1.05& 2.49 & 1.18 & 2.59   &  6.13   & 19 \\ 
    \hline
    \textbf{Kinem.‐based}        & 3.89/2.23 & 5.37/2.60  & 3.00/1.85  & 7.85/4.32  &2.93/1.52   &   8.22/4.00   & 1.08/ 1.04&
    0.3/1\\ 
    \hline
    \end{tabular}
    \label{tab:OL}
    \vspace{-1.0em} 
\end{table*}

We present our results in Tables \ref{tab:OL} and \ref{tab:cl}, which evaluate the transition from open-loop benchmarks to closed-loop driving performance in a realistic simulation.
Table~\ref{tab:OL} reports standard OL metrics on the validation set (ID), the entire nuPlan validation set (OOD), and predictions at each timestep during closed-loop simulation in the selected highly interactive scenarios (SIM).
In addition, we assess average inference time $t_i^{SIM}$ and the temporal consistency of the predictions defining the following metric, which is averaged over all samples: \begin{equation} TC^{SIM} = \frac{|\hat{x}^m_{i,t|N} - \hat{x}^m_{i,t+1|N-1}|}{v_{i,t}} \end{equation}
This metric quantifies the change in predictions between consecutive timesteps, normalized by the current velocity.

Subsequently, Table \ref{tab:cl} summarizes the CL performance of different predictor-planner combinations in the interactive simulation scenarios introduced in Sec.~\ref{sec:exp} based on the metrics introduced in Sec.~\ref{sec:related_work}. In the following, we discuss the key observations and insights drawn from these experiments.

\textbf{OL ID Evaluation to OL In-Simulation Evaluation:}
Table \ref{tab:OL} first highlights that training on a union of diverse datasets as facilitated by UniTraj appears to increase robustness in out-of-distribution (OOD) tests (e.g. \textit{minADE}$^\mathrm{ID}_6$ vs. \textit{minADE}$^\mathrm{OOD}_6$), with much smaller performance drop than reported in prior work (\eg, \cite{yao_improving_2024, unitraj}). 
Interestingly, MTR-Mini and Wayformer-Mini maintain performance levels closer to their original counterparts in OOD settings, suggesting potential overfitting of their larger counterparts.
Despite only minor performance drops in OOD tests on the nuPlan dataset, we observe a notable decrease when evaluating the same predictors on the predictions made during simulation rollouts (i.e., for each timestep of the nuPlan simulation in combination with a planner). This suggests a distribution shift that remains significant. We attribute this to the more challenging nature of the selected interactive scenarios. Hence, where the prediction quality is most important from safety perspective, we see a high performance drop due to long-tail events. Additionally, the planner’s feedback on the vehicle's behavior and interplay with the predictor in these environments may further accentuate this shift.
\begin{table*}[!htb]
    \vspace{0.5em}
    \caption{CL performance of predictors combined with MPCC  and RBMPCC, evaluated across (left to right): speed, absolute jerk, minimum TTC per scenario, reporting average(+standard deviation) and collision rate, and nuPlan score.}
    \vspace{-0.6em}
    \centering
    \begin{tabular}{|p{2.05cm}||m{1.0cm}|m{1.25cm}|m{1.4cm}|m{0.8cm}|m{0.9cm}||m{1.0cm}|m{1.25cm}|m{1.4cm}|m{0.8cm}|m{0.9cm}|}
    \hline
    \rule{0pt}{9pt}
     & \multicolumn{5}{c||}{\textbf{MPCC}} & \multicolumn{5}{c|}{\textbf{RBMPCC}} \\
    \hline

     & $ v$[m/s]$\uparrow$
     & $|j|$[m/s³]$\downarrow$
     & $TTC_{\min}$[s]$\uparrow$
     & $CR$ $\downarrow$
     & \textbf{Score} $\uparrow$
     & $ v$[m/s]$\uparrow$
     & $|j|$[m/s³]$\downarrow$
     & $TTC_{\min}$[s]$\uparrow$
     & $CR$$\downarrow$
     & \textbf{Score}$\uparrow$ \\
    \hline
    \textbf{Oracle}
      & 9.2 (4.4) & 0.19 (0.31) & 0.84 (0.82) & 12\,\% & 0.746
      & 8.6 (4.1) & 0.22 (0.37) & 1.11 (0.79) & 2\,\%  & 0.881 \\
    \hline
    \textbf{MTR}
      & 8.9 (4.3) & 0.20 (0.38) & 1.10 (0.91) & 12\,\% & 0.697
      & 8.4 (3.8) & 0.28 (0.37) & 1.20 (0.96) & 8\,\%  & 0.818 \\
    \hline
    \textbf{MTR-Mini}
      & 9.2 (4.2) & 0.20 (0.31) & 0.84 (0.87) & 16\,\% & 0.697
      & 8.2 (3.9) & 0.27 (0.36) & 1.33 (0.87) & 2\,\%  & 0.850 \\
    \hline
    \textbf{Wayformer}
      & 9.4 (4.3) & 0.20 (0.34) & 0.90 (0.80) & 14\,\% & 0.654
      & 8.2 (3.9) & 0.26 (0.37) & 1.28 (0.87) & 6\,\%  & 0.837 \\
    \hline
    \textbf{Wayformer-Mini}
      & 9.4 (4.3) & 0.18 (0.33) & 0.99 (0.81) & 10\,\% & 0.693
      & 8.3 (3.9) & 0.26 (0.34) & 1.14 (0.77) & 4\,\%  & 0.836 \\
    \hline
    \textbf{AutoBot}
      & 9.5 (4.4) & 0.21 (0.38) & 0.91 (0.80) & 14\,\% & 0.585
      & 8.2 (4.0) & 0.26 (0.36) & 1.23 (0.90) & 8\,\%  & 0.790 \\
    \hline
    \textbf{AutoBot-Mini}
      & 9.6 (4.4) & 0.24 (0.56) & 0.73 (0.71) & 14\,\% & 0.580
      & 8.1 (4.1) & 0.27 (0.34) & 1.36 (0.88) & 4\,\%  & 0.805 \\
    \hline
    \textbf{Kinematic-based}
      & 8.9 (4.5) & 0.21 (0.42) & 0.93 (0.89) & 18\,\% & 0.697
      & 7.9 (3.6) & 0.30 (0.42) & 1.02 (0.70) & 4\,\%  & 0.748 \\
    \hline
    \end{tabular}
    \label{tab:cl}
    \vspace{-1.0em}
\end{table*}

\begin{figure}[!htb]
\centering
    \includegraphics[trim = 6mm 0mm 8mm 8mm, clip, width=0.49\textwidth]{"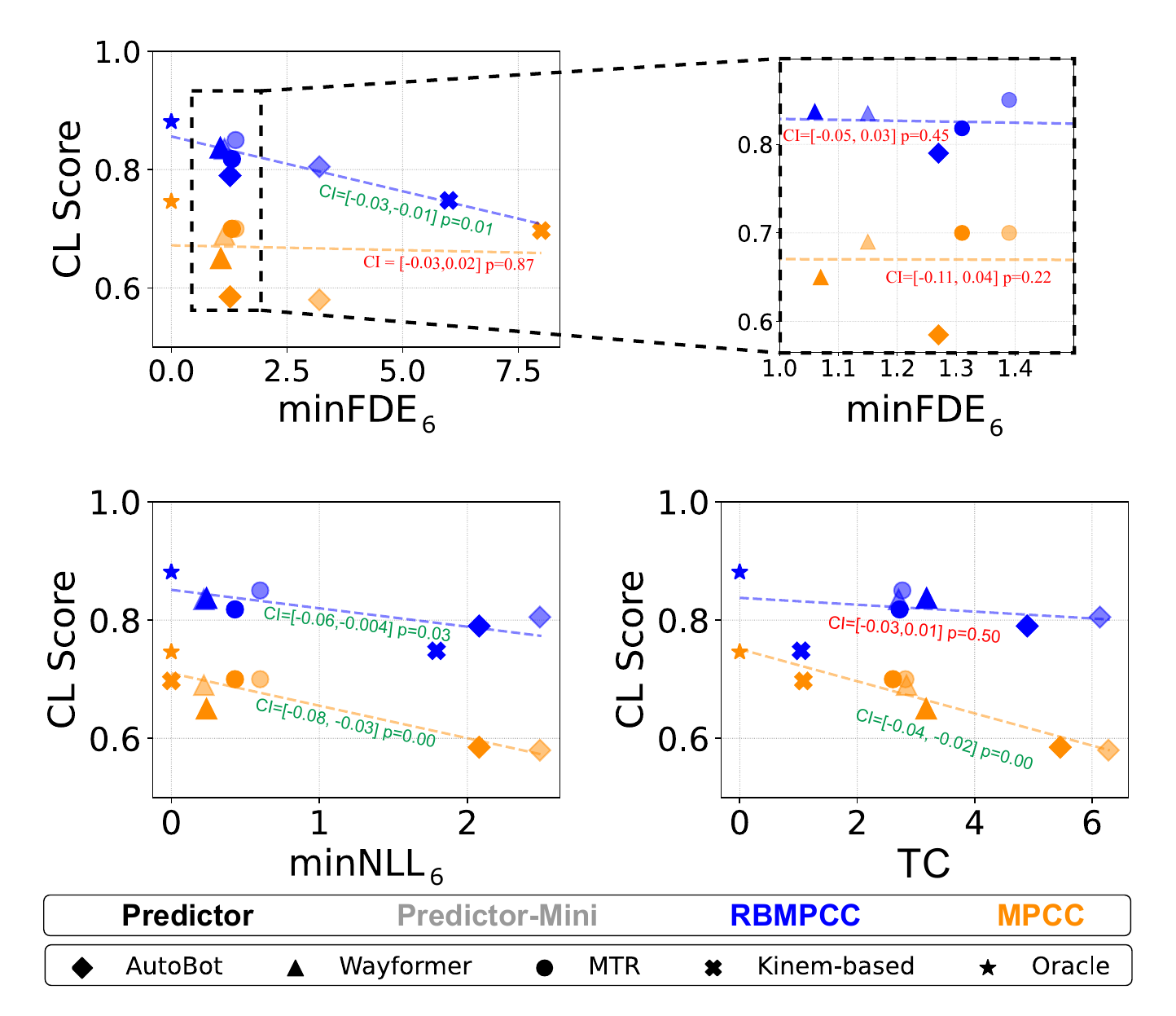"}
    \vspace{-1.3em} 
    \caption{OL scores vs. CL average score over all scenarios for each predictor-planner combination. The top-right panel shows a zoomed-in view of the top-left plot. minFDE and minNLL are obtained from ID tests, while temporal consistency is computed as the average across all simulation runs. A t-test on the regression slope coefficient is performed under the null hypothesis, \ie, slope equals zero. A slope is considered statistically significant if the corresponding p-value is less than 0.05. For each regression, we report the confidence interval (CI) and the p-value, for the zoomed-in plot we repeat that with only the learning-based predictors.}
    \label{fig:P1} 
    \vspace{-0.4em} 
\end{figure}

\begin{figure}[!htb]
\centering
    \includegraphics[trim = 0mm 0mm 0mm 0mm, clip, width=0.49\textwidth]{"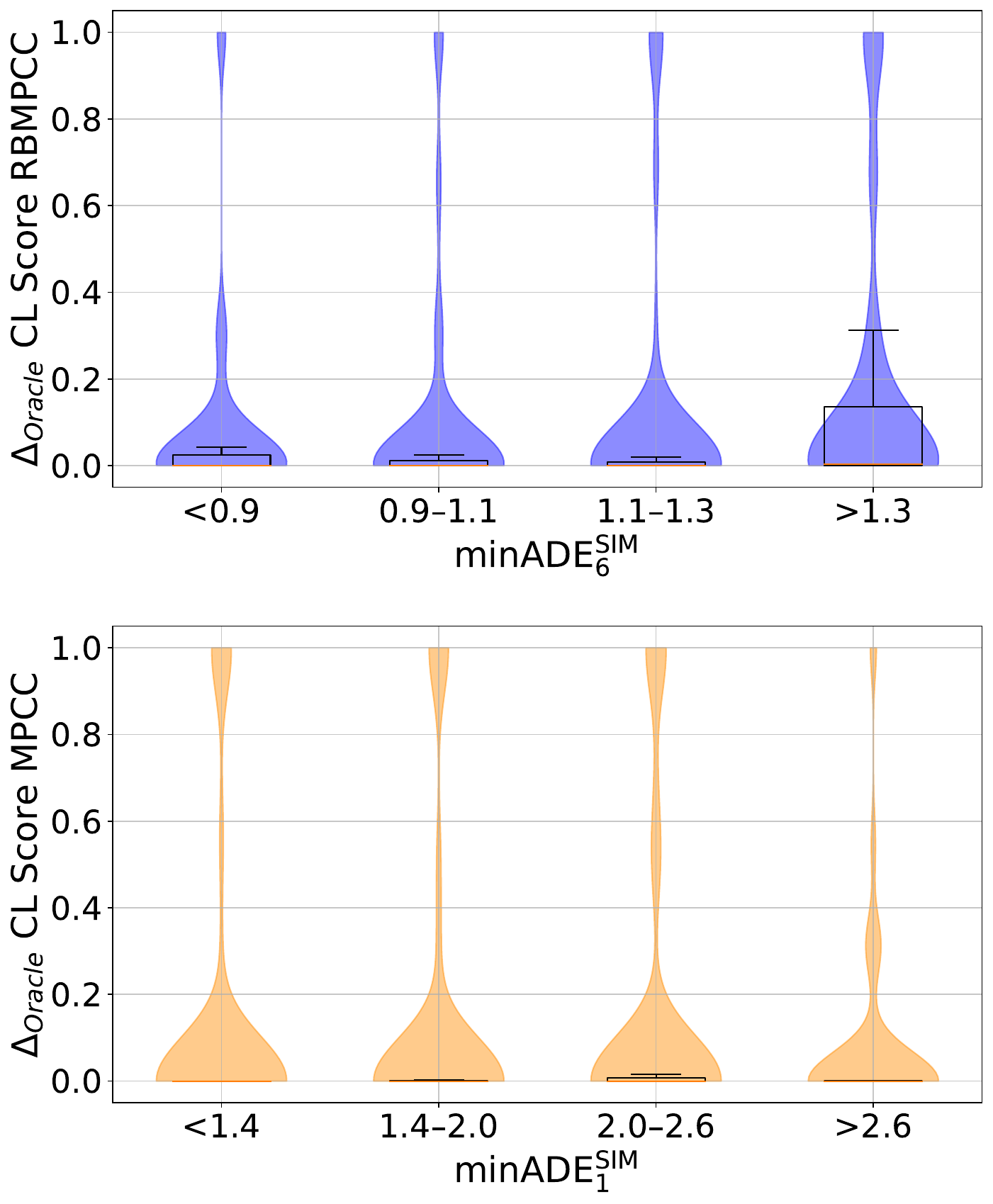"}
    \vspace{-1.3em} 
    \caption{ Illustration of planner performance change based on the predictor's OL performance. We show the difference in CL score relative to the oracle predictor as a function of each simulation’s minADE) . For each scenario–predictor pair, we compute the minADE over the full simulation, bin the results into four intervals with approximately equal numbers of scenarios, and display the distribution of $\Delta_{Oracle}$CL (CL score minus oracle CL score) within each bin. For MPCC runs, we use the minADE of the most probable mode ($K=1$), as only this mode is used for planning. In rare cases where the predictor outperforms the oracle, we cap $\Delta$CL at zero (i.e. zero means no more potential for improvement) to avoid distorting the distribution.}
    \label{fig:P2} 
    \vspace{-0.4em} 
\end{figure}

\begin{figure}[!htb]
\centering
    \includegraphics[trim = 12mm 0mm 8mm 0mm, clip, width=0.49\textwidth]{"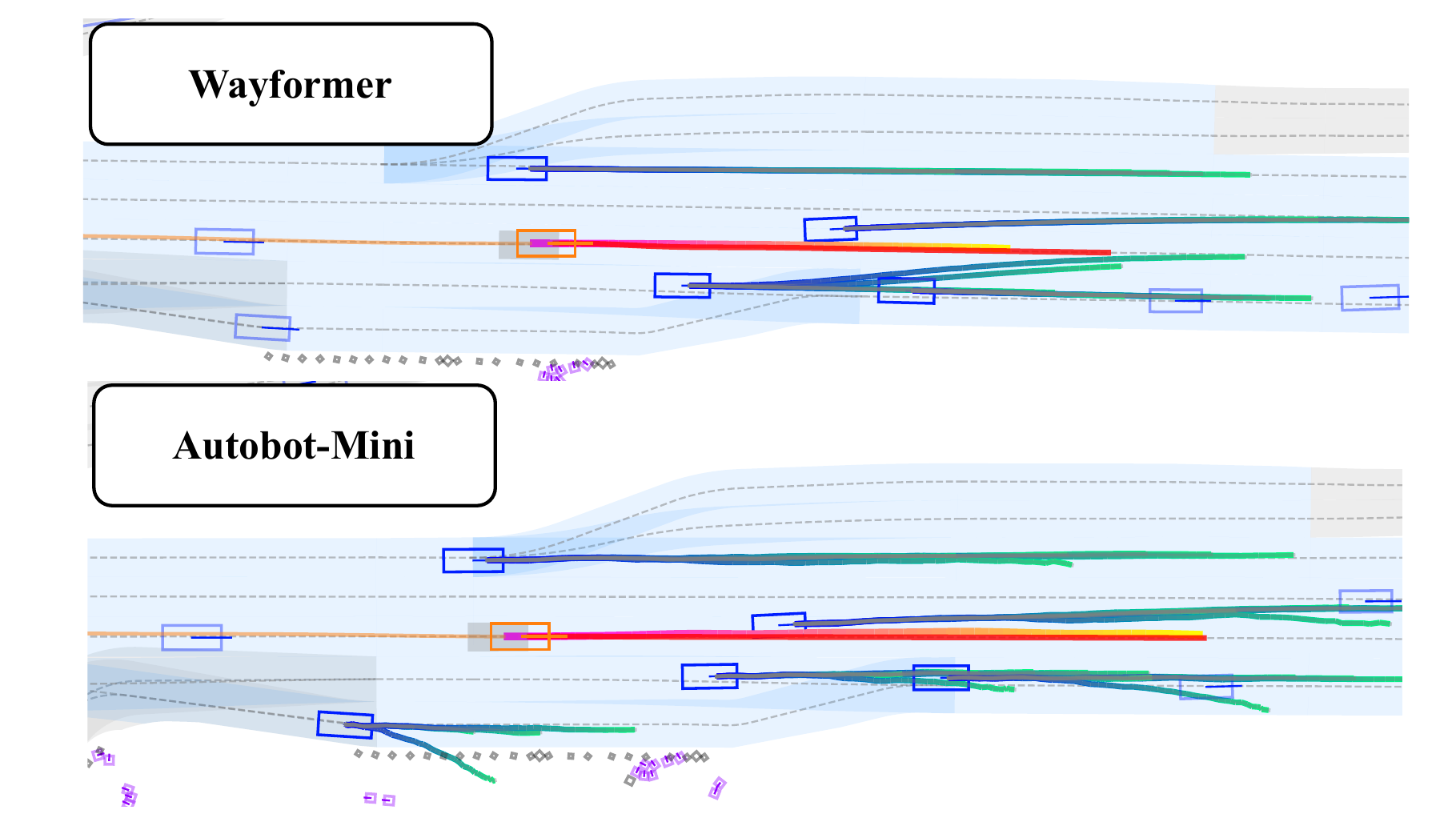"}
    \vspace{-1.3em} 
    \caption{An example for the relevance of diversity in the prediction. The agent with Wayformer drives more carefully and has a contingency plan (yellow) to brake in case the traffic participant changes the lane.}
    \label{fig:P4} 
    \vspace{-0.4em} 
\end{figure}

\textbf{Different planner - different predictor requirements:}
We find that different planners respond differently to predictor accuracy (cf. Fig \ref{fig:P1}, \ref{fig:P2}). MPCC exhibits relatively stable performance across predictors except for Autobot models, which degrade sharply. Here the kinematic predictor has the best performance. In contrast, RBMPCC performs worst with the kinematic predictors and the learning-based models performing best. This suggests that only sufficiently sophisticated planners can fully exploit the complexity of advanced predictors.
The limited benefit of multi-modal predictors with MPCC likely stems from its planning with only a single mode. RBMPCC, on the other hand, can leverage the uncertainty inherent in multi-modal predictions through contingency planning (Fig \ref{fig:P4}). This results in higher Time-To-Collision (TTC) and lower collision rates, as well as a substantial overall performance boost.

\begin{figure}[!htb]
\centering
    \includegraphics[trim = 0mm 0mm 0mm 0mm, clip, width=0.49\textwidth]{"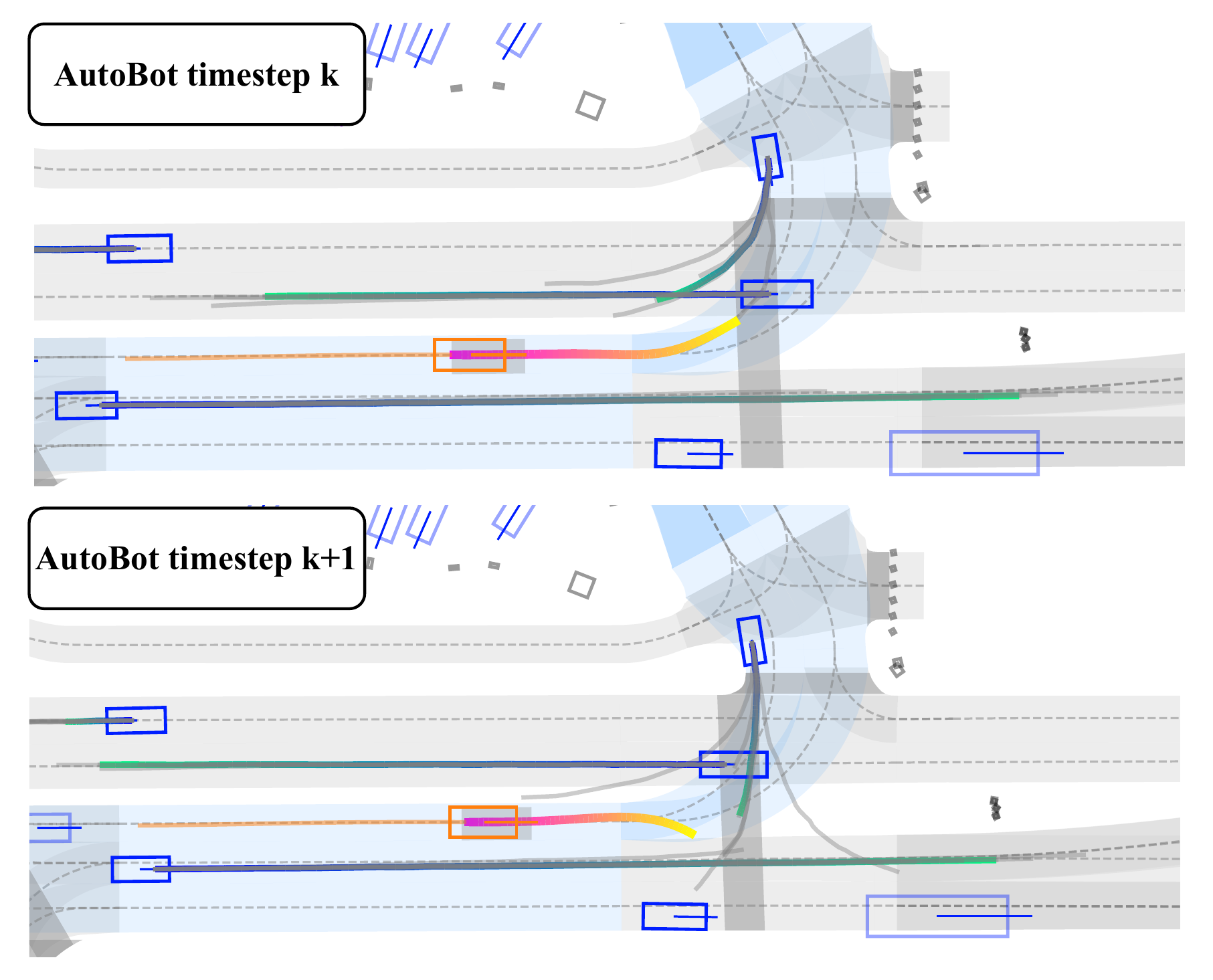"}
    \vspace{-1.3em} 
    \caption{ An example for the relevance of temporal consistency in the prediction. The predictors and thus the planner in this example jump between different behaviors which leads to jerky driving style and convergence problems of the planner.}
    \label{fig:P3} 
    \vspace{-0.4em} 
\end{figure}

Fig. \ref{fig:P1} shows that temporal consistency plays a more critical role for MPCC. This could explain, why the two Autobot models exhibit severe performance degradation with MPCC, likely due to instability in their predicted modes leading to the worst temporal consistency score by far.  When predictions fluctuate over time, MPCC is forced to replan frequently, leading to jerky behavior and poor convergence (Fig. \ref{fig:P3}). This instability is also linked to their high minNLL, indicating low confidence in the most likely predicted mode.

Fig. \ref{fig:P1} also indicates that minNLL has a noticeable impact on MPCC performance. 
The Autobot models exhibit the highest minNLL values by a significant margin. When the predicted probabilities of different modes are close to each other, the ranking of the mode may fluctuate across timesteps, potentially explaining the poor temporal consistency observed in Autobot's predictions.

RBMPCC, in contrast, appears less sensitive to temporal consistency and minNLL. Its contingency-based planning strategy may even benefit from prediction uncertainty, using the resulting diversity of future trajectories to make more cautious, robust decisions (Fig \ref{fig:P4}).

\textbf{Accuracy-related OL Metrics vs. CL performance:}
Comparing predictor rankings in Table \ref{tab:OL} (open-loop) and in Table \ref{tab:cl} (closed-loop) reveals substantial shifts. For instance, MTR-Mini ranks 5th in open-loop but becomes the best-performing model in closed-loop driving, particularly when paired with RBMPCC.
This indicates that beyond a certain threshold, further improvements in predictor accuracy, as measured by open-loop metrics, do not necessarily translate into better closed-loop performance. This trend holds both for in-distribution (ID) accuracy (Fig. \ref{fig:P1}) and for predictions within simulation (Fig. \ref{fig:P2}). In the case of RBMPCC, performance degradation only becomes apparent at very low-accuracy models (e.g., with kinematic predictors), and for the MPCC even there no performance drop is visible 

Fig. \ref{fig:P2} shows that, with RBMPCC, closed-loop driving remains robust up to 1.3 m ADE in simulation. In contrast, MPCC does not appear to benefit meaningfully from improved ADE.

\textbf{Predictors model size  reduction vs. CL performance:}
While reducing model size leads to some degradation in open-loop ID performance, closed-loop performance remains largely unaffected. In fact, MTR-Mini consistently outperforms its larger counterpart in closed-loop driving. These results support the hypothesis that larger models may overfit to open-loop benchmarks achieving higher rankings without improving real-world driving outcomes.

This finding is especially relevant for real-world deployment, where hardware and latency constraints often require compact models. In our experiments, the inference time for MTR dropped significantly with model size, while other models saw only minor improvements. This discrepancy is due to the parallelization ability of hardware accelerators. Once resource limits are reached, additional FLOPs (cf. (Tab. \ref{tab:splits})) start to dominate computation time. In our case, only MTR saturated the hardware (NVIDIA A10G GPU, 32 vCPUs, 128 GiB RAM), but on embedded systems or production vehicles with more limited compute, these effects would be more pronounced.

\section{Conclusion}\label{sec:conclusion}
We introduced a framework to evaluate different motion prediction models in the context of downstream planning performance. Our results demonstrate that improvements in OL metrics such as minADE or minFDE do not always translate to better closed-loop behavior. This highlights a critical misalignment between current benchmarking practices and real-world autonomous driving performance.
By systematically analyzing different combinations of SotA predictors and planners, we revealed that the effectiveness of a predictor depends not only on its standalone accuracy but also on how well it interacts with the planner. Sophisticated planners like RBMPCC can exploit multimodal predictions, while simpler ones like MPCC may benefit more from stable, deterministic inputs.
Surprisingly, we found that downsized models, despite lower OL accuracy, can match or even exceed the driving performance of their larger counterparts in closed-loop evaluations. This has important implications for model efficiency and deployability on resource-constrained platforms.
Our findings emphasize the need for system-level evaluation metrics and more nuanced benchmarking strategies that consider the full prediction-planning stack.

\bibliographystyle{IEEEtran}
\bibliography{references}
 
\end{document}